\documentclass[conference]{IEEEtran}
\IEEEoverridecommandlockouts
\usepackage{cite}
\usepackage{amsmath,amssymb,amsfonts}
\usepackage{algorithmic}
\usepackage{graphicx,caption}
\usepackage{textcomp}
\usepackage{xcolor}
\usepackage{hyperref}
\usepackage[font=footnotesize]{caption}
\usepackage{subfig}

\def\BibTeX{{\rm B\kern-.05em{\sc i\kern-.025em b}\kern-.08em
    T\kern-.1667em\lower.7ex\hbox{E}\kern-.125emX}}
    
\definecolor{babyblue}{rgb}{0.54, 0.81, 0.94}
    
\begin{document}

\title{Real-Time Oil Leakage Detection on Aftermarket Motorcycle Damping System with Convolutional Neural Networks
}

\author{\IEEEauthorblockN{1\textsuperscript{st} Federico Bianchi}
\IEEEauthorblockA{\textit{Idea-re S.r.l. } \\
Perugia, Italy \\
fbianchi@idea-re.eu}
\and
\IEEEauthorblockN{2\textsuperscript{nd} Stefano Speziali}
\IEEEauthorblockA{\textit{Idea-re S.r.l. } \\
Perugia, Italy \\
sspeziali@idea-re.eu}
\and
\IEEEauthorblockN{3\textsuperscript{rd} Andrea Marini}
\IEEEauthorblockA{\textit{Idea-re S.r.l. } \\
Perugia, Italy \\
amarini@idea-re.eu}
\and
\IEEEauthorblockN{4\textsuperscript{th} Massimiliano Proietti}
\IEEEauthorblockA{\textit{Idea-re S.r.l. } \\
Perugia, Italy  \\
mproietti@idea-re.eu}
\and
\IEEEauthorblockN{5\textsuperscript{th} Lorenzo Menculini}
\IEEEauthorblockA{\textit{Idea-re S.r.l. } \\
Perugia, Italy \\
lmenculini@idea-re.eu}
\and
\IEEEauthorblockN{6\textsuperscript{th} Alberto Garinei}
\IEEEauthorblockA{\textit{Department of Engineering Sciences,} \\
\textit{ Guglielmo Marconi University}\\
\textit{\& Idea-re S.r.l. }\\
Perugia, Italy \\
a.garinei@unimarconi.it}
\and
\IEEEauthorblockN{7\textsuperscript{th} Gabriele Bellani}
\IEEEauthorblockA{\textit{Department of Industrial Engineering,} \\
\textit{Alma Mater Studiorum Università di Bologna }\\
Bologna, Italy \\
gabriele.bellani2@unibo.it}
\and
\IEEEauthorblockN{8\textsuperscript{th} Marcello Marconi}
\IEEEauthorblockA{\textit{Department of Engineering Sciences,} \\
\textit{ Guglielmo Marconi University}\\
\textit{\& Idea-re S.r.l. }\\
Perugia, Italy \\
m.marconi@unimarconi.it}
}

\maketitle

\begin{abstract}
In this work, we describe in detail how Deep Learning and Computer Vision can help to detect fault events of the AirTender system, an aftermarket motorcycle damping system component.  One of the most effective ways to monitor the AirTender functioning is to look for oil stains on its surface. Starting from real-time images, AirTender is first detected in the motorbike suspension system, simulated indoor, and then,
 a binary classifier determines whether AirTender is spilling oil or not. The detection is made with the help of the Yolo5 architecture, whereas the classification is carried out with the help of a suitably designed Convolutional Neural Network, OilNet40.  In order to detect oil leaks more clearly, we dilute the oil in AirTender with a fluorescent dye with an 
 excitation wavelength peak of approximately 390 nm.  AirTender is then illuminated with suitable UV LEDs.  The whole system is an attempt to design a low-cost detection setup. An on-board device, such as a mini-computer, is placed near the suspension system and connected to a full hd camera framing AirTender.  The on-board device, through our Neural Network algorithm,  is then able to localize and classify AirTender as normally functioning (non-leak image) or anomaly (leak image).
\end{abstract}

\begin{IEEEkeywords}
Sensors, Computer Vision,  Convolutional Neural Networks, AirTender, Fluorescence.
\end{IEEEkeywords}

\section{Introduction}

Nowadays, automotive diagnostics plays a significant role in ensuring vehicle driving safety and comfort. In particular, to guarantee the driver's safety, it is crucial to monitor constantly the chassis system components, such as the damping apparatus.  The most immediate and coarse methods to check the health state of the damping system are provided by the driver's perception and, more  generally, human inspection.
  Other more quantitative approaches for fault detection and isolation are classified as reliability-based, model-based, signal-based or statistical-based. See \cite{kothamasu2006system} for details.

Having an automated system that takes care of the health state of a vehicle is important also because of the increasing demand of public services such as shared mobility or car rentals where human inspection is not always possible.  Traditionally, automotive diagnostics has often relied on ``on-board diagnostics'', where sensors allow generation of data within a vehicle which are then processed for diagnostic purposes.  See \cite{siegel2017survey} for a survey on connected vehicles.

Sensors can record signals, such as vibrations or noise,  and algorithms suitably designed should process such data to benchmark the current status of the vehicle against standard conditions previously defined \cite{delvecchio2018vibro}.  This approach has many advantages and, in general,  its purpose is to help detect and isolate fault events.  When failures are not properly detected by on-board diagnostics (or are detected with delay), one could also consider off-board diagnostics \cite{siegel2020surveying}, 
which is interesting in its own right.
 We will not be exploring off-board diagnostics in this paper. Rather, we
  focus our attention on 
 an
  on-board diagnostic system suitably designed to detect oil leakages from a special type of motorcycle suspension system introduced shortly.

An outstanding paradigm for detecting faults is that of Machine Learning (ML), where data coming from sensors can be fed to algorithms able to detect (with a certain degree of certainty) fallacies or anomalies.  Deep Learning (DL) in machine health monitoring \cite{zhao2019deep} often concerns the use of ML architectures such as Auto-Encoders, Restricted Boltzmann machines, Convolutional Neural Networks,
 and Recurrent Neural Networks.  They all provide outstanding models according to the task at hand. We  give more details and references for how DL is used in automotive diagnostics in the next section.

As mentioned above, visual inspection or the driver's perception are the first parameters taken into account for the vehicle monitoring. However, it is often the case that perception is not enough to detect minor fault events.  A more systematic analysis of the suspension system behavior, based on data collected during normal functioning,  may help considerably in this case. A first-class approach consists in feeding Neural Network
 signals coming from sensors installed on the vehicle. In particular, Convolutional Neural Networks (CNNs) turn out to be good feature-extractors and quite robust under changing conditions \cite{zhao2016research}.  However,  vibration analysis usually relies on the physical installation of suitable accelerometers on the vehicle. Such sensors may sometimes not be available or have an elevated cost.  Moreover, in the case of suspensions such as
 those studied in this work, one of the main fault precursors is oil spilling, which usually occurs even before a major break takes place. 

In this work, we focus our attention on oil leak
 detection in an aftermarket suspension 
system---AirTender---via Computer Vision and CNNs.  AirTender is a shock absorbing system produced by UmbriaKinetics (for more details on AirTender,
 visit the website \url{https://airtender.it}, accessed on 17 October 2022). 
In short, AirTender is a system that combines the advantages of the traditional coil spring system with that of the air suspensions present in today's motocross. The body of AirTender constitutes the core of the system: it is made of two annular elements (cylinders) sliding one inside the other and forming an oil chamber, pressurized by the compressed air present in the external cylinder. 

The strategy we follow for leak detection is rather simple and can in principle be applied to other suspension systems as well. Our goal is to determine whether a motorbike suspension is experiencing oil leakage close to the main damper by exploiting UV oil fluorescence.  As an aside, oil leakage detection from various types of equipment and starting from oil fluorescence has been already widely discussed in the literature. 
See, for example \cite{chase2010real, camagni1988diagnostics,  fingas2014review, sato1978method}. 
 For our purposes, we set up two Convolutional Neural Networks (CNNs) working in sequence, one that localizes the core of AirTender in the suspension system by drawing a bounding box around it, and the second one which classifies the content of the bounding box as a leak/non-leak AirTender.

From a practical point of view, the idea is to place an on-board device such as a mini-computer near the suspension system. A full hd camera connected to the on-board device frames AirTender and the surrounding environment (tarmac, wheels, vehicle body,
 etc.).  The on-board device, through our Neural Network algorithm,  should then be able to detect/localize AirTender. After AirTender has been detected correctly,  it undergoes a classification process as normally functioning (non-leak) or anomaly (leak) and, in case of an 
 anomaly, informs the driver of the fault.

The setup is rather easy to put together and is illustrated in the main body of the paper.  The final goal is to have an automatic detection system,  part of the on-board diagnostics, able to spot oil leaks on the suspension system.  As this is a first study, all tests have been made on the 
test bench showing promising results.

This paper is organized as follows.  In Section \ref{related work}, we briefly describe the broad domain of vehicle diagnostics and the problem of detecting suspension system failure.  We also give a short review of ML-based fault diagnosis  methods.  Section \ref{Proposed approach} is the core section of this paper. We first briefly summarize the basics of Convolutional Neural Networks, with focus on their application in computer vision.  Then,
 we present the experimental setup and its main features along with the details of how we prepared our own dataset, needed to train our algorithm to spot oil leaks.  Finally, we lay out the method we used to train our algorithm and we spell out in detail the network architectures we employed.  In \linebreak Sections \ref{results} and \ref{conclusions},
  we give the results and conclusions of our analysis.

\section{Related Work}\label{related work}

Over the years, system health monitoring has been part of the set of activities performed on a system to keep it in good working condition \cite{kothamasu2006system}. 

System monitoring can either be limited to the observation of the current status of a system or can be assisted by predictions of future operating states as well as future failure
 states. In the former case, maintenance comes to play after an observation suggests it, while in the latter case,
  it can also play  a preventive role. 

In general, predictive diagnosis and prognosis are motivated by the need for manufacturers to optimize system performance and reduce extra costs due to unwanted disruptions.  Thus, it does not come as a surprise that system health monitoring has been the focus of research in the 
recent years.

In this paper, our goal is to discuss an approach to monitor the health of a suspension system.  Standard approaches in monitoring the health state of the damping system of a vehicle are often model-based \cite{ferreira2009sensing, hernandez2015fault,  hernandez2013fault} or signal-based \cite{alcantara2016fault}.  A model-based approach usually relies on the comparison between the measured signal (e.g., periodic vibration signals captured by the accelerometers) and the estimated values generated by a mathematical model of the system, while a signal-based approach focuses on detecting changes or variations in a signal with subsequent diagnosis of the change.

However, one crucial problem in automotive diagnosis is robustness with respect to vehicle configuration and changing conditions (mass variations, different tire characteristics,
 etc.), usually captured as driving data.  As a matter of fact, model-based and signal-based approaches do not immediately take advantage of such data, if not with a fine-tuning method. 
  Therefore, in the last few years, data-driven approaches leveraging deep learning models have come to play a central role in current research \cite{zhao2019deep}. 

As mentioned above, when dealing with machine health monitoring systems,  it is customary to use signals coming from sensors, such as accelerometers or potentiometers and give them as input to Neural Networks.  Some authors operate directly on raw-data (i.e., no pre-processing on measured signals), while others implement some transformation, such as
 scaling, denoising, Fourier transformation,
  etc. 

For example,  in \cite{xia2017fault},
 a CNN consisting of two convolutional and pooling layers followed by a dense layer was trained on raw data coming from multiple sensors.  In addition,
  the authors of \cite{eren2019generic} trained
   a CNN consisting of three convolutional layers and two sub-sampling layers followed by a dense layer taking directly raw time-series sensor data as input.  In \cite{zilong2018intelligent}, the authors combined the concept of ``inception'' and one-dimensional deep convolutional neural networks to train a CNN to operate directly on raw vibration signals.

On the other hand,  the authors of \cite{janssens2016convolutional} performed
 a frequency analysis before feeding data to a CNN architecture.  Several types of bearing faults, along with a healthy bearing and rotor imbalance, were considered. For each condition, several bearings were tested.  

Another common approach is to create grayscale images (2D) by reshaping time signal vectors to 2D tensors. For example, the authors of \cite{wen2017new} proposed
 a new CNN, based on LeNet-5 \cite{lecun1998gradient}, for fault diagnosis.  They were able to extract features of converted 2D images and eliminate the effect of handcrafted features. The proposed method is tested on three famous datasets, including the 
 motor bearing dataset, self-priming centrifugal pump dataset, axial piston hydraulic pump dataset.  
 In addition,
  \cite{zhang2017bearings} performed preprocessing of 1D temporal vibration signals by generating 2D grayscale images.  Then, classification was carried out by exploiting a CNN made of two convolutional layers followed by a pooling layer. 

Other preprocessing methods that have become standard in health monitoring in Machine Learning comprise Wavelet Transform (WT), Short-Time Fourier Transform (STFT) or Hilbert--Huang Transform (HHT).  See, for example  \cite{liao2017wavelet,  verstraete2017deep, zhang2018fault, ding2017energy}, where CNN architectures are employed on preprocessed time series data.   

It is worth noting at this point that, in \cite{zehelein2020diagnosing}, CNNs played a major role also for the diagnosis of automotive semi-active suspension components. The authors fed a CNN (described in the paper) both raw and preprocessed data coming from seven different sensors (four wheel speeds, lateral and longitudinal accelerations,
 and yaw rate).  Inputs are given as both 1D and 2D tensors. 

Even though CNNs applied to (preprocessed) actual driving data promise to give excellent results, our approach is different.  As brought forward in the Introduction, the aim of the present study is to set up a convenient ``visual system'' able to detect fault events (through oil leakages) with a camera framing the suspension system. To this end, we have CNNs processing actual images of the suspension system described in detail in the next~section.

\section{Proposed Approach}\label{Proposed approach}

In this section,
 we describe the proposed approach for oil leak detection. This is the longest section of the paper.  We first give an overview of CNN architectures and how they work in general. Then,
  we describe our experimental setup and dataset and finish off by laying out the details of the method we used for classification and the CNNs we employed for the analysis of this paper.

\subsection{Overview of CNNs}\label{CNNs}

Convolutional Neural Networks are known to be good feature
 extractors, especially when dealing with multidimensional data as inputs. 
 They came to popularity as a method to classify images, and the first working prototype was proposed by LeCun \cite{lecun1989handwritten} for image processing. A comprehensive overview of CNN lies outside the the scope of the present article. The reader who is interested in learning the nuts and bolts of CNNs is referred to the excellent books \cite{geron2019hands, chollet2021deep} or the review \cite{mehta2019high} and references therein.  For a more formal treatment of CNNs and deep learning in general, see \cite{bronstein2017geometric,  bronstein2021geometric}. 
 Very useful is also the thesis~\cite{cohen2021equivariant}. Here, we
   just summarize their main features and how they can be useful for our purposes. The reader that already knows CNNs and their application to Computer Vision can safely skip this part.

\subsubsection{Structure of CNNs}

The building blocks of Convolutional Neural Networks are \textit{convolutional} and \textit{pooling layers}. 
A convolutional layer computes the convolution (a mathematical operation that we define shortly) of the input image with a stack of filters.  Downsampling or pooling layers, instead, coarse-grain the input preserving locality and spatial structures. 

In Machine Learning, it is customary to think of images as functions on a two-dimensional compact domain,  $f: \Omega \rightarrow \mathbb{R}^{C}$, where $C$ is often referred to as the number of channels and is equal to 1 for grayscale images, while it is equal to 3 for color images. Typical examples of compact domains are the Euclidean rectangle $\Omega = W \times H$ or $\mathbb{S}^2$ for spherical CNNs. Here, $W$ stands for the width and $H$ for the height of the input image (in practice, Euclidean domains are always discretized by means of a regular grid). 
Such a construction naturally suggests a real vector bundle structure for input images, where the base space is the domain $\Omega$ and the fiber is simply $\mathbb{R}$ for grayscale images or $\mathbb{R}^3$ for color images (here, the group of symmetry transformations on images usually corresponds to a subgroup of $\mathrm{GL}(2, \mathbb{R})$ or the translation group in two dimensions.  Elements of the fibers usually belong to the trivial representation. In addition,
 the bundle is often trivial \cite{cohen2021equivariant}).  
Function images are often required to be square-integrable over $\Omega$, and therefore belong to $f \in L^2(\Omega)$.  In a computer vision setting, we seek for an unknown function $y: L^2(\Omega) \rightarrow \mathcal{Y}$ defined on a training set of images $\{ f_i \in L^2(\Omega)\}$. For instance, when dealing with image classification, $\mathcal{Y}$ can be thought of as a discrete set with $\text{rank}(\mathcal{Y})$ the number of classes, whereas if dealing with object localization,
 $\mathcal{Y}$ is a multi-dimensional simplex.  

Let us give a little more mathematical detail about convolution and pooling, as promised. Here,
 we heavily draw from \cite{bronstein2017geometric,  bronstein2021geometric}. A convolutional layer is a layer of the form $g = C_{\Gamma}(f)$. Essentially, we act on a $p$-dimensional input image $f(x)$ with a bank of filters $\Gamma = (\gamma_{k,k'})$, producing a $q$-dimensional output $g$,
\begin{equation}
g_{k}(x) = \xi \left( \sum_{k' = 1}^{p} f_{k'} \star \gamma_{k, k'} (x) \right) \, .
\end{equation}
Here, $k=1, \dots, q$ and $k' = 1, \dots, p$ while $\xi$ is a nonlinear pointwise activation function, such as the often employed ReLU activation function, $\xi(x) = \text{max}(0, x)$ \cite{nair2010rectified}.  The output $g$ is usually called \textit{feature map},  which, roughly highlights those areas of the input that most activate the filter $(\gamma)$.  The convolution operation is defined for a planar CNN as
\begin{equation}
(f_{k'} \star \gamma_{k,k'})(x) = \int_{\Omega} \text{d}x' \gamma_{k,k'}(x') f_{k'}(x - x') \, .
\end{equation}

The filters $\gamma$ correspond to the \textit{shared weights}. In a convolutional network, these weights correspond to the free parameters that are tuned by the learning algorithm in order for the network to extract the most important features of the input image.  This allows the network to 
focus on---and learn---the important features of the input image.

Besides convolutional layers, we have pooling layers.  They correspond, roughly, to layers which reduce the size of feature maps, decreasing, in turn, also the model parameters without losing too much information about the input. Their goal is to sub-sample the input image in order to reduce the computational burden. Aside from that, they also introduce to some level a notion of \textit{translational invariance}, one of the hallmarks of CNNs.   Pooling layers can be defined as
\begin{equation}
g_{k} (x) = P \{ f_{k}(x') : x' \in \mathcal{N}(x)\} \, , \quad k = 1, \dots, q \, ,
\end{equation}
where $\mathcal{N}(x)$ is a neighborhood of $x$ in $\Omega$ and $P$ is a permutation-invariant function. Common pooling operations are \textit{max pooling} or \textit{average pooling} \cite{nagi2011max, lin2013network}. 

A convolutional network is then put together by assembling a stack of convolutional and pooling layers, 
%$U_{\Theta}(f) = (C_{\Gamma^K} \dots P \dots C_{\Gamma^1})(f)$,
\begin{equation}
U_{\Theta}(f) = (C_{\Gamma^K} \dots P \dots C_{\Gamma^1})(f) \, ,
\end{equation}
where $\Theta = \{ \Gamma^1  \dots \Gamma^K  \}$ is the vector of parameters of the network (shared weights) (pooling layers are not characterized by learning parameters).

In a supervised setting,  the CNN parameters can be learned by asking for the minimization of some loss function measuring the ``distance'' between the outputs $(U_{\Theta}(f_i)$'s and the labels $y_i$'s, 
%$\mathcal{L} = \sum_{i \in \mathcal{I}} L (U_{\Theta}(f_i), y_i)$.
\begin{equation}
\mathcal{L} = \sum_{i \in \mathcal{I}} L (U_{\Theta}(f_i), y_i) \, .
\end{equation}
If the model is complex enough and the dataset representative, the algorithm is expected to generalize outside the training set \cite{abu2012learning}. 

Usually,  convolutional and pooling layers are followed by dense (fully connected) layers and a classifying activation function, such as
 softmax or sigmoid, 
 depending on the goal of the CNN.  We 
  employ precisely such a structure (con-pool layers followed by dense layers) later in the paper when  classifying the leak/non-leak status of AirTender.  More modern architectures, especially when performing segmentation or object detection, contemplate the use of Fully Convolutional Networks (FCNs) \cite{long2015fully}. 

In the present paper, we  mainly deal with two CNN-type architectures, the Yolov5~\cite{jocher2020yolov5} (introduced later) for object localization and a suitably designed CNN (referred to in the following as OilNet40) created to classify the status of AirTender. 
 Full details are given in the following sections.

We would like to stress at this point that CNNs have not only been successful to process images, but have also been used to address sequential data, including Natural Language Processing and speech recognition. See,
 e.g., \cite{geron2019hands} for pedagogical details and further references.  However, given the goal of this paper, we 
  focus more on their power to process physical images.

\subsubsection{CNNs for Computer Vision}

Computer Vision deals with how computers can derive high-level meaningful information from digital images or videos.  The synergy with Deep Learning, through DL architectures such as
 CNNs, has produced dramatic results in recent years.  This is essentially due to the application of CNNs through 2010 to 2017 to the ImageNet Large Scale Visual Recognition Challenge, or ILSVRC \cite{ILSVRC15}.  Tasks of ILSVRC usually comprise \textit{image classification}, \textit{image localization},
  and \textit{object detection}. 

In computer vision, classification usually refers to predicting the classes of objects present in an image. The difference between object localization and detection is a little subtle.  Object localization aims at locating an object in an image, while object detection tries to find out all the objects and their boundaries. In this paper, the terms localization and detection are
 used interchangeably as they define a common task in our algorithm, even though this is not standard practice.  

Over the last few years,
 different convolutional CNN architectures have been proposed. Among these, those that came to prominence for their degree of accuracy are LeNet-5 \cite{lecun1998gradient},  AlexNet \cite{krizhevsky2012imagenet},  GoogLeNet \cite{szegedy2015going},  VGGNet \cite{simonyan2014very}, ResNet \cite{he2016deep}, Inception-v4 \cite{szegedy2017inception}, Xception \cite{chollet2017xception}, GBD-Net \cite{zeng2017crafting}, SENet \cite{hu2018squeeze}. They constitute a milestone in the field of Deep Learning applied to Computer Vision.  For a detailed account on the genesis of all these architectures and their structure see, for example, \cite{geron2019hands}. By leveraging CNN architectures such as
  those just introduced, people were able to design Neural Networks for object detection as well. Famous deep learning architectures for object detection are: Yolo \cite{redmon2016you,  redmon2017yolo9000, redmon2018yolov3}, SSD \cite{liu2016ssd},  Faster R-CNN \cite{ren2015faster},  Mask R-CNN \cite{he2017mask}. For a detailed account on the different object detection architectures, see \cite{zhao2019object}. In this paper,
   we  use a Yolo architecture for
    our object detection task.  

\subsubsection{CNNs and Transfer Learning}\label{transfer learning}

It is often the case that training a deep neural network takes a huge amount of time and computational power, and very few companies have such considerable resources. Moreover, it is often wise to store the knowledge gained, so that we re-use a trained model or  build on that model.  This is in essence what \textit{transfer learning} is for. 

Many famous CNN models, such as 
AlexNet, GoogLeNet, ResNet, InceptionNet, VGGNet, etc., have been released as part of useful packages such as the Torch Vision library in Pytorch or Keras in Tensorflow.  These models can be loaded directly to perform inference on the same (or similar task) they were trained for, or they could even be used as baseline for training in related problems.  This is in essence what we
 do when training the Yolov5 to detect AirTender: we
  load the Yolov5s network with pre-trained weights and train it to identify AirTender.  On the other hand, we are
   not using transfer learning for the classification task, as random weight initialization is
    sufficient. 

The next section is related to the experimental setup.

\subsection{Experimental Setup}\label{setup}

The problem with oil spilled on a opaque surface is that it is oftentimes hardly visible, even to a human eye.  To make it more easily detectable, we employed a method often used to spot oil leaks: observation of emitted fluorescence when the oil is irradiated with ultraviolet (UV) light.  In order to amplify the UV fluorescence,
 we also diluted the oil with fluorescent dye.  Following the specifications on the mixing of the dye (max percentage of 0.7$\%$ of dye), oil mechanical features are not compromised. This,  in fact, turned out to be a crucial step to make the oil more clearly visible by the camera and, in turn, more detectable.

We also assembled a simple setup with 4 UV LEDs emitting light at of 385 nm along with other 5 UV LEDs emitting at 400 nm wavelength.
  The LEDs we used emit directional light, with an opening angle of 15 degrees.  
  They were powered with a 12 V battery, consuming 0.6 W.

The detection of fluorescence emitted by oil leaks irradiated with UV light has been discussed before in the literature. See, for example, \cite{lu2019oil} (and references therein) for oil leak detection with solar irradiation and image processing using deep learning. It is often the case that observing fluorescence in conditions of high illumination is not an easy task, essentially because the noise produced by daytime light makes fluorescence hardly visible. However,  AirTender is hardly ever exposed to direct sunlight when fully assembled on a motorcycle. Thus, we reckon that in condition of moderate illumination (50--2000 lux), oil fluorescence from UV light should be clearly visible.

\begin{figure}[h]
	\centering
	\includegraphics[width=5cm]{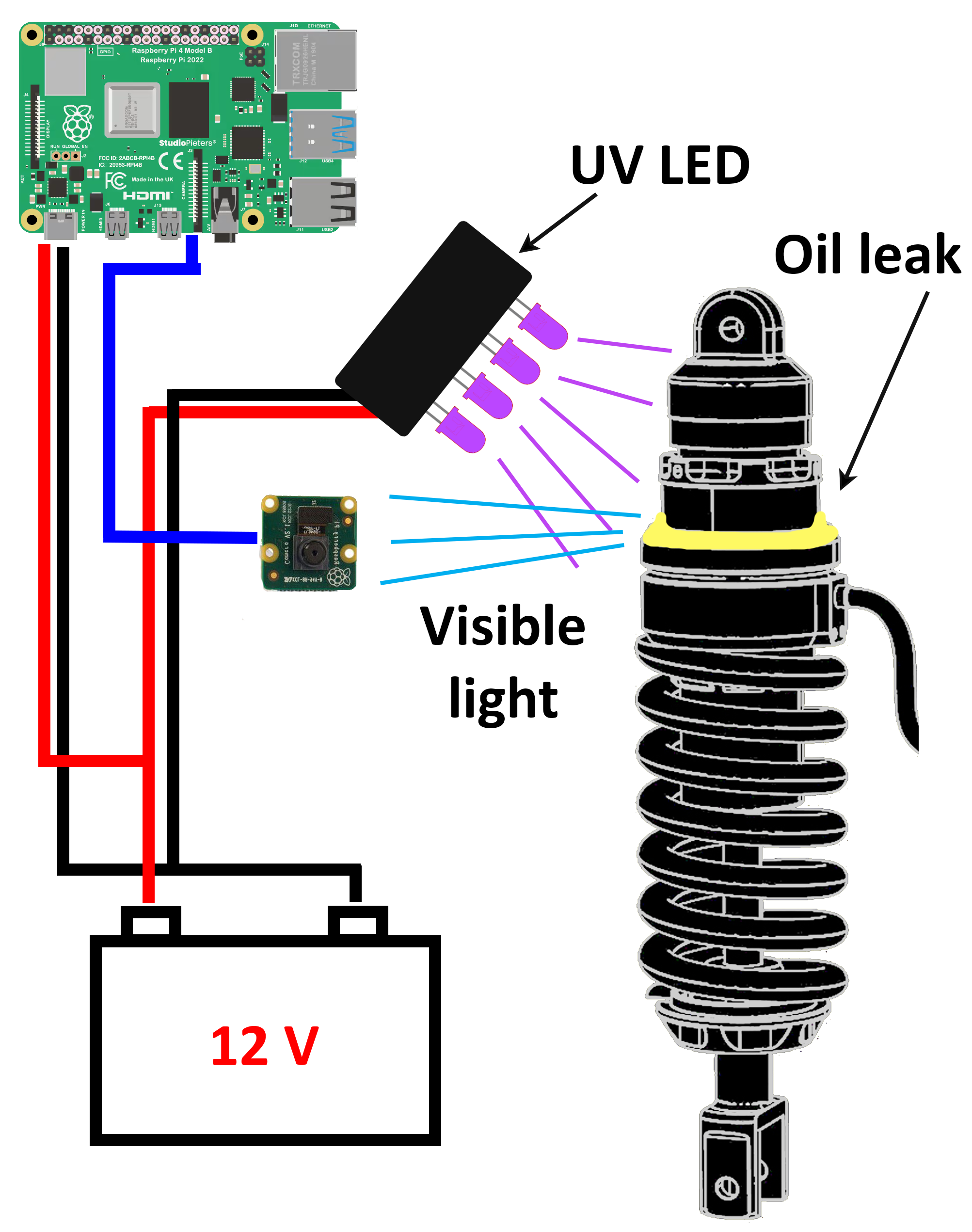}
	\caption{Setup to find oil leaks on the AirTender suspension system.  A camera connected to an on-board device records the core of AirTender, while UV LEDs make oil fluoresce.  The LEDs and the on-board device are powered by the on-board battery. The on-board device elaborates the images recorded by the camera through the neural networks trained for this study.}
	\label{fig:ATandfluorescence}
\end{figure}

The remaining part of the experimental setup consists of an OrangePi 3 mini-computer, with a 64-bit 1.8 GHz ARM processor, a multi-core GPU Mali T720 and 2GB of RAM. A full-HD camera is connected via USB to the OrangePi. The UV LED box is connected indirectly to the 12V battery via a relay, driven by the OrangePi. When collecting photos, the relay is turned on and the suspension system is enlightened with UV light (Figure~\ref{fig:ATandfluorescence}). The OrangePi runs a python-based algorithm to detect and classify the AirTender, exploiting the TensorFlow Lite runtime, a library specifically designed to fit small computers, as mobiles and edge computing devices (see \url{https://www.tensorflow.org/lite},  accessed on 17 October 2022,  for full details on TF Lite). Detection and classification algorithms elaborate real-time images
 acquired by the camera.

We now proceed to discussing how we created the dataset to train our networks to distinguish a leaking AirTender from a non-leaking one.

\subsection{Dataset}\label{dataset}

In this section, we give some details on how the dataset for AirTender detection/classification was created. 

As is well-known, CNNs require a large number of images for their training.  
In addition, 
the dataset should be as diverse as possible, in order for the CNN to generalize well outside the training set. At the same time, in a supervised learning setting, such images require labels. Producing labels for computer vision tasks such as object detection, classification or even segmentation can easily become challenging and time consuming.  In these cases,
 techniques such as transfer learning or data augmentation can come to the 
 rescue, and simplify the process of creating a representative dataset.  We discussed briefly what transfer learning is in Section \ref{transfer learning}, while we  touch on data augmentation  in Section~\ref{data augmentation}.

We made our own dataset for AirTender detection as well as for classification (leak or non-leak images). 
 In an
  object localization task, labels correspond to the bounding boxes around the object we want to locate, while in a binary classification task such as ours, one would simply associate one of the two classes to each image, leak or non-leak in the present~case.

We generated the dataset by extracting photograms from a number of videos of AirTender as well as by taking pictures of AirTender directly.  The collected images are aimed at representing AirTender in different conditions: different motorcycle setups, different backgrounds as well as different conditions of lighting, orientations,
 etc.  More importantly, some of the collected images should simulate oil leakages on AirTender.

Eventually, we collected 1634 different images which were divided into a training, validation and test set for the training. 

We ended up having a training set made of 1044 images, a validation set made of 413 images,
 and a test set made of 177 images.  All images have dimension 1300 $\times$ 1300 $\times$ 3, meaning that they are color images. In addition,
   they correspond to images taken in condition of illumination in the range 50--2000 lux.  As this is a first study, we collected our dataset trying to represent AirTender in as many different conditions as possible, but of course, a more refined dataset could be taken by considering even more different setups, different image quality,
    and so on. We leave the creation of a larger dataset for the future, limiting ourselves to having a fully working algorithm.

Labels for object detection can be created using standard tools such as LabelImg, see,
 e.g., \cite{tzutalin2015labelimg}. Alternatively, one can rely on web User Interfaces such as that provided by Roboflow (\url{https://docs.roboflow.com}, accessed on 17 October 2022). 
(The script we used is available at \url{https://github.com/ideare-ds/tools\textunderscore for\textunderscore Computer\textunderscore Vision}, accessed on 17 October 2022.). 
When producing labels for a Yolov5 network,
 we have to make sure to save the labels (one for each image) as a text file (.txt). 
 The annotation file  consists
  of a string of five numbers. The first number identifies the class the object to detect belongs to. The remaining four numbers identify the coordinates of the center of the bounding box ($x$ and $y$) as well as the width and height of the box.  All numbers are normalized to lie between 0 and 1. 

Notice that we aim at building a classifier that distinguishes a leaking AirTender from a normally functioning one. Thus, we should prepare a dataset with a balanced number of images of Air Tender on full display with either spilled oil on it (faulty AirTender) or just a normal Air Tender with no oil leakage and thus working well. Indeed, when training a classifier,
 leak or non-leak images are associated to two different classes.

Let us move on to discussing the algorithm methodology to classify leak/non-leak~images.

\subsection{Method}\label{method}

In this section, we spell out the details of our method to detect oil leaks on AirTender. We first lay out the steps we followed to train our algorithm and then discuss three methods to improve training: image preprocessing, data normalization,
 and data augmentation.

\subsubsection{Workflow}

To determine whether the AirTender is leaking or not, we want to propose a two-step process.  We first spatially localize the AirTender in the suspension system of the morcycle: a camera placed near the suspension system will,  in general,  record images of AirTender and its background and, thus, it is important to first detect and isolate AirTender from its surrounding.  Then,
 the resulting detected images are
  passed through a classifier which labels leak or non-leak images.

The steps we followed to have a fully working algorithm for oil leak detection are as follows.  We start with our dataset, the creation of which has been discussed in the previous section.  We first apply some data augmentation (see Section \ref{data augmentation} for details) in order to have a larger and more diverse dataset.  Images are then passed through the Yolov5 (training) which should localize AirTender within the image by enclosing it in a bounding~box.

The content of the bounding box (supposedly AirTender) is then cropped and preprocessed. The resulting image passes trough the binary classifier (OilNet40) which outputs the leak/non-leak status of AirTender.

\subsubsection{Image Preprocessing}\label{image preprocessing}

In the previous sections, we discussed a method to amplify oil fluorescence. We might wonder whether there is more we can do to make oil leakages even more visible. It turns out that some image preprocessing can be helpful to highlight the spilled oil, especially if we could improve contrast to highlight oil fluorescence.

In this work, we relied on the application of Contrast Limited Adaptive Histogram Equalization (CLAHE) to input images. CLAHE is a variant of the Adaptive Histogram Equalization (AHE) which prevents over-amplification of the contrast, and operates on small regions of the image, also called tiles, rather than the entire image. 

It turns out that histogram equalization techniques, such as CLAHE, are implemented in OpenCV \cite{opencv_library}.  It is possible to choose parameters such as the clip limit, i.e.,  the value at which the histogram is clipped, and 
 the tile size. 

In order to apply CLAHE to all input images, we first convert RGB images to HSV (Hue, Saturation, Value) space. We then apply CLAHE to the channel H (Hue) and convert back to RGB colors. This helps highlight fluorescence of oil leaks making them more visible.

\begin{figure*}[h!]
\centering
\captionsetup{width=.42\linewidth}
\subfloat[Normally functioning AirTender processed in four different ways: original image (A), CLAHE filter (B), gray + CLAHE (C) and CLAHE + gray (D)]
{\includegraphics[width=6.9cm]{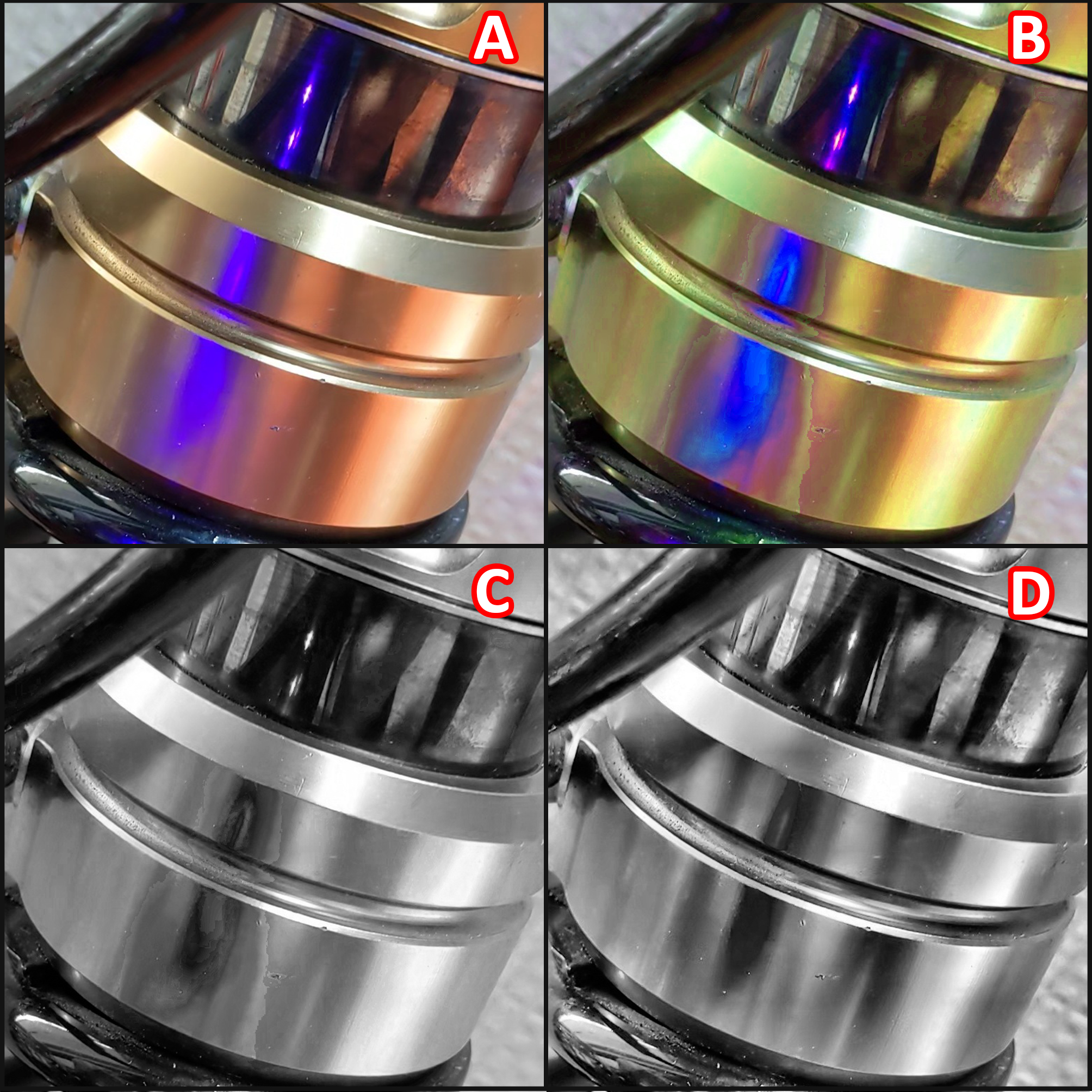}
\label{acc}}
\hfil
\subfloat[Faulty AirTender: original image (A), CLAHE filter (B), gray + CLAHE (C) and CLAHE + gray (D). Oil fluorescence is
	 clearly visible on the core of AirTender.]
{\includegraphics[width=6.9cm]{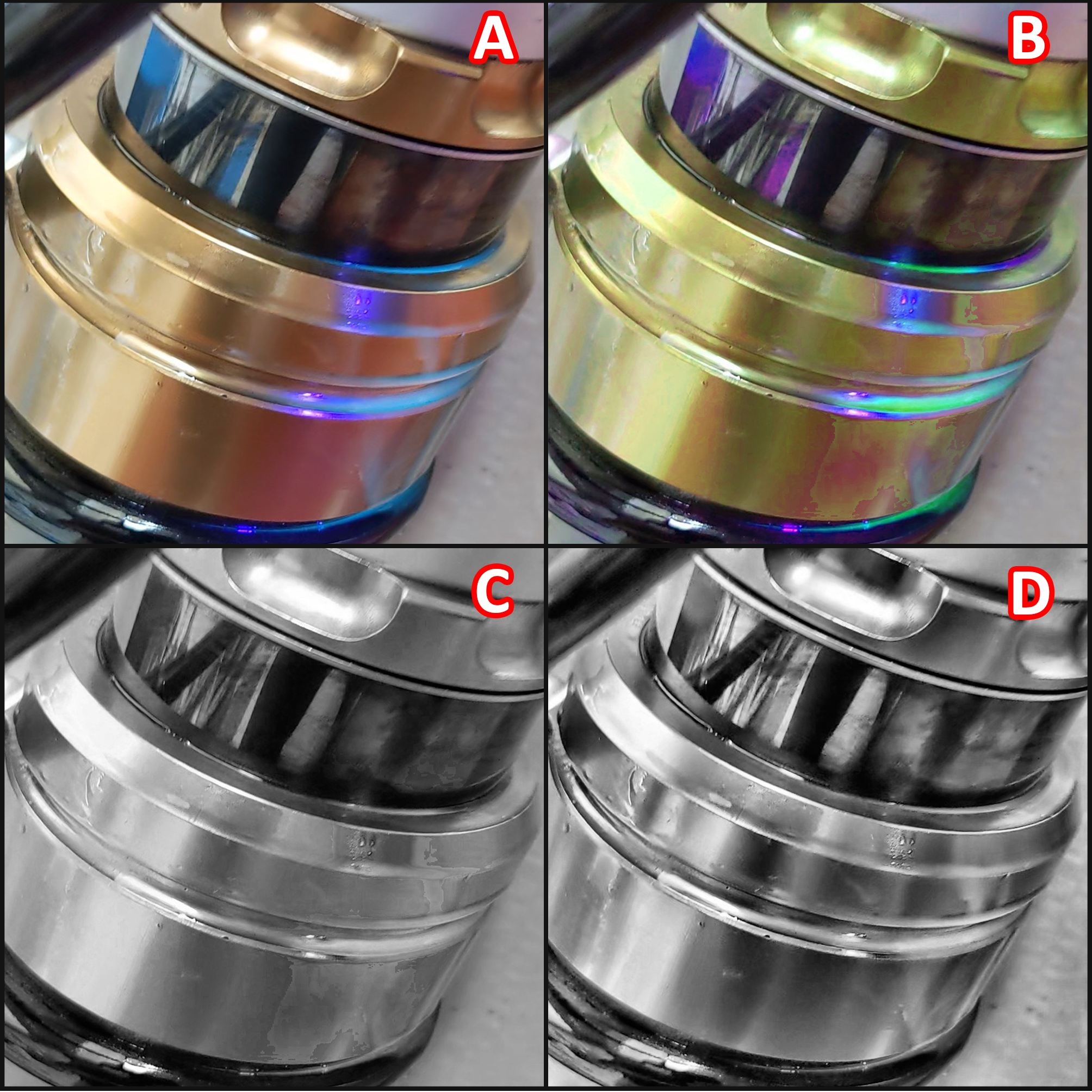}
\label{prec_rec}}
\caption{}
\label{fig:four_datasets}
\end{figure*}

Besides color images, we also applied CLAHE before and after converting color images (RGB channels) to grayscale (1 channel) images, producing two different datasets for a total of 4 different presentations of the same dataset. See Figure~\ref{fig:four_datasets} for the four different preprocessing methods applied to a normally functioning AirTender (left) and a faulty AirTender (right). Eventually, we test our classification algorithm on all datasets. 

\subsubsection{Image Normalization}

For most images,  the pixel values are usually given in terms of integers in the range 0 to 255.  On the other hand, neural networks process input data using small weights and biases. It is often the case that neural networks work better when the pixel values are comparable with weights and biases.  This, in general, helps the algorithm converge faster. 

Thus, a step often taken in computer vision is to normalize the input data, rescaling the pixel values to lie in the range 0 to 1. This is readily done by dividing each pixel of each color channel by 255.

\subsubsection{Data Augmentation}\label{data augmentation}

Data augmentation is a useful technique in Machine Learning used to increase the size of the training set. This is particularly useful when creating a fully representative dataset. 
It turns out to be a challenging task.  In addition,  CNNs by construction leverage translation invariance and scale separation (as explained above), but do not explicitly take into account different kinds of transformations that do not spoil information about the image.  Examples of such transformations are rotations, dilations, different conditions of lighting, and so on. Thus, data augmentation can be a useful tool to manually include such a transformation without changing the network architecture.

For this work, we relied heavily on data augmentation for both tasks, AirTender localization and classification.  In more detail, we considered random rotation of input images,  random zoom, shifts and flips both vertically and horizontally as well as random change of brightness conditions.  The Yolov5 architecture also contemplates a mosaic method for data augmentation, where random images are put together (mosaic) and, therefore, more objects of the same type can appear in the same input image.

\subsection{Model Architectures}\label{models}

In this section, we summarize the model architectures we employed for detection and classification, respectively. We begin with a brief summary of the Yolov5 architecture and the setup we used for the training. Then, we move on to discussing our architecture for AirTender classification.

\subsubsection{Model Architecture for AirTender Detection: Yolov5}

For object detection, we relied on the Yolov5 architecture \cite{jocher2020yolov5}.  Yolov5 is a family of object detection architectures pretrained on the COCO dataset and based on the Pytorch framework \cite{NEURIPS2019_9015}. Full details on the model architecture as well as tutorials on how to train the network on new datasets and deploy the model can be found on GitHub \cite{jocher2020yolov5}.

Within the Yolov5 family, there are 10 different architectures, pretrained on the COCO dataset, which differ in size and performance.  As explained on the GitHub page, a Yolov5 model can be exported to different formats, including TensorFlow SavedModel, TensorFlow Lite,
 and TensorFlow.js (and many others). This is a useful feature if one wanted to deploy a model to a desktop or a web app. 

For our algorithm, we used the Yolov5s architecture, a light version with less parameters than its siblings Yolov5m, Yolov5l or Yolov5x but with a shorter inference time.  This choice was mainly dictated by the fact that we want to deploy the trained model to single-board computers such as OrangePi or RaspberryPi, which should be able to detect AirTender in real time.

\subsubsection{CNN Model for Binary Classification: OilNet40}
\label{subsec:oilnet}

In this paper, we use our own deep CNN called OilNet40. The network was developed using Keras \cite{chollet2015keras}, the high-level API of TensorFlow 2 \cite{tensorflow2015-whitepaper}.  The purpose of OilNet40 is to perform binary classification on images of AirTender. In other words, it should distinguish between leak and non-leak images of AirTender.  Such a network was designed following standard rules of building CNNs, see,
 for example,
  \cite{chollet2021deep, geron2019hands}.  The reason for designing a specific CNN, rather than using some well-established model and leveraging transfer learning, lies in 
   the ability of its customization: the setup of very complex networks performing convoluted and diverse tasks is less prone to modifications and is less malleable to our purposes.

\begin{figure}[h]
	\centering
	\includegraphics[width=8.3cm]{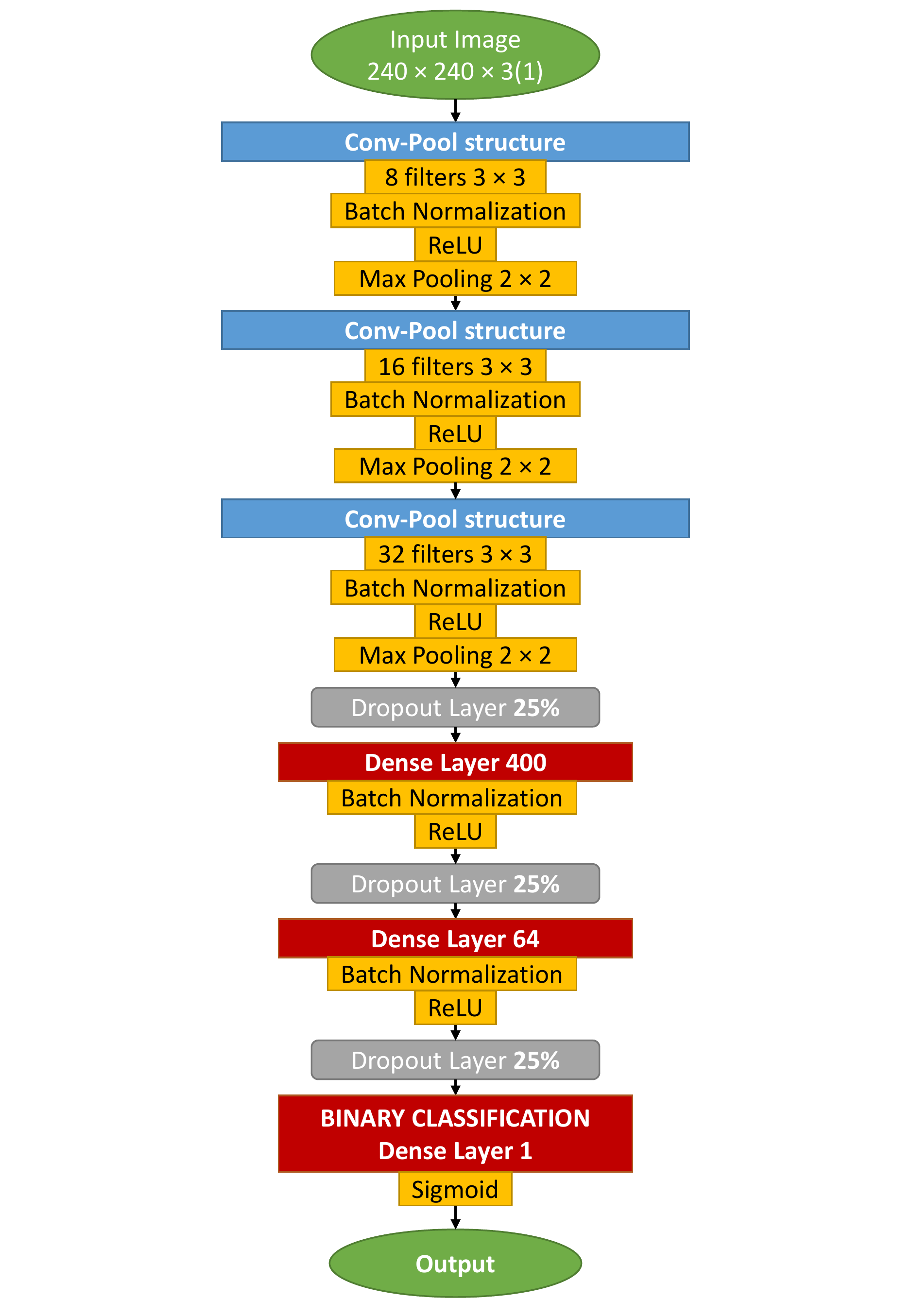}
	\caption{OilNet40 CNN diagram architecture.}
	\label{fig:oilnet40}
\end{figure}

The structure of the network is shown in Figure~\ref{fig:oilnet40}.  Input images are first resized to 240 $\times$ 240 $\times$ 3 and then passed through a stack of convolutional and dense layers. 

In particular, we have three stacks of convolutional layers, with 8, 16,
 and 32 $3 \times 3$ filters, respectively.  Convolutional layers are all followed by batch-normalization \cite{ioffe2015batch} and ReLu activation function \cite{nair2010rectified}.  Here, batch normalization is used after the convolution.  The ReLu activation function introduces nonlinearity to the network and is one of the most commonly used activation functions. Convolution is always performed with $3 \times 3$ filters.  Some authors consider also filters with different dimensions. In general, the dimension of filters is seen decreasing along the neural network. It is not uncommon to use 5 $\times$ 5 or even 7 $\times$ 7 sized filters, for the first convolutional layers for large input images.  Finally, $2 \times 2$ max-pooling is considered after each convolutional layer. 

After convolutions and max-pooling, we included a stack of two dense layers.  Indeed, little experimentation showed that one stack of dense layers is not enough to process the features extracted with convolutional filters, resulting in worse classification. On the other hand, three stacks of neurons introduce
 too many parameters, at the risk of overfitting.  We introduced a 
 dropout ($25 \%$) before each dense layer. Dropout is a regularization technique often used to reduce overfitting. It selects randomly neurons to be removed during each training, reducing in turn the number of parameters of the network. 
  The first dense layer has 400 neurons, while the second one has 64 neurons. Each dense layer is followed by batch normalization and a ReLu activation function.  At the top of the neural network,
   we have a single neuron where a sigmoid gives back the prediction of the network, i.e., leak or non-leak Airtender with 0.5 as threshold.

For the training, we used the Nadam optimizer, i.e., Adam with the Nesterov trick~\cite{dozat2016incorporating}. In addition,
 we tested different CNN architectures, changing each time the number of filters, number of neurons in the dense layers and learning rate of the optimizer. All this was done with the help of Keras Tuner \cite{omalley2019kerastuner} Random Search. In particular, we set 
 up a grid search in order to tune the following parameters: number of neurons in the dense layers and the learning rate. Finally, we selected the model performing best on the validation set. Of all the architectures,
  we explored that
  the best was the 40th network in the list, hence the name~OilNet40.

\section{Results}\label{results}

In this section, we collect the results of our classification algorithm. As explained before, we trained two different neural networks, the Yolov5s for the detection of AirTender and the OilNet40 network for the actual classification.

The classification is performed on four different datasets, each corresponding to a different preprocessing technique. In particular, the first dataset is the original dataset, with no preprocessing. In the second dataset, images are processed with a CLAHE filter, while the third and fourth correspond to grayscale images obtained by conversion to grayscale before and after applying the CLAHE filter, respectively (see Section \ref{image preprocessing}). Results for the training of OilNet40 on all four datasets are presented below.

\subsection{AirTender Detection}

We begin by summarizing the results for the detection of AirTender. To this end, we used the Yolov5s architecture \cite{jocher2020yolov5}.  As anticipated, we trained the small Yolov5 architecture because the model eventually is to be deployed to our mobile system,  and a network with a relatively small number of parameters is easier to export and use, besides offering faster~performance. 

As mentioned above,  the dataset  comprises
  1634 images, split into a training, a validation,
  and a test set.  The dataset we work with is not made of a very large number of images. This is essentially due to the many technical difficulties that arise when producing one. Nonetheless, we tried to make it as diverse as possible, using images of AirTender in different conditions of lighting,  orientation, positioning,
   etc.  In addition,
    we relied heavily on data augmentation, as offered by the Yolov5 system \cite{jocher2020yolov5}.  This helped us work with an even more diverse dataset.  In particular,  random rotation in the range $-$90 to 90 degrees along with vertical and horizontal flipping allows to have in the dataset images of AirTender framed from all different angles. This is particularly useful when, in the final setup, one does not have much control on the relative orientation of the recording camera and AirTender. Moreover, the mosaic data augmentation feature allows to have images of AirTender at different distances from the recording camera. 

\begin{figure}
	\centering
 	\includegraphics[width=8.3cm]{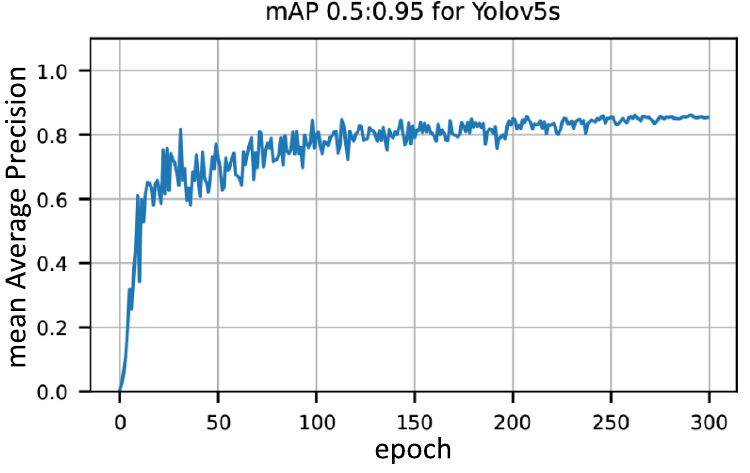}
   	\hspace{-0.5cm}\caption{Mean Average Precision (mAP 0.5:0.95) for AirTender detection.}
   	\label{fig:yolo_metrics}
\end{figure}

In Figure~\ref{fig:yolo_metrics},  we show the result of the training. In particular, we show the mean Average Precision mAP 0.5:0.95 against the number of epochs. 
It is used in many benchmark challenges such as Pascal, VOC or COCO, and corresponds to a number normalized between 0 and 1. Roughly, the closer it is  to 1, the easier it is for the system to detect the AirTender at the correct position.  See, e.g., \cite{geron2019hands} for details.

The actual training was performed on a PC equipped with a GPU Nvidia RTX 3060 with 12 GB of dedicated memory and 3584 CUDA cores.  We see in Figure~\ref{fig:yolo_metrics} that the model is learning throughout and achieves a mAP of about 85$\%$ only after 300 epochs.  

Overall, we find that the Yolov5 architecture is rather easy to train and works reasonably well, at least in the case presented in this paper.  Once AirTender is correctly localized, its status can be classified by a second network. Thus, we can now proceed to discussing the more interesting part of the algorithm, i.e., classification of leak or non-leak images of AirTender.

\subsection{AirTender Classification}

Classification is the second task of our algorithm. Once the AirTender is correctly located in the suspension system, the aim is to spot oil leakages, if any. To this purpose,  as anticipated before, we set up a convolutional network (OilNet40) for classification.

\begin{figure}[h]
	\centering
	\includegraphics[width=8.3cm]{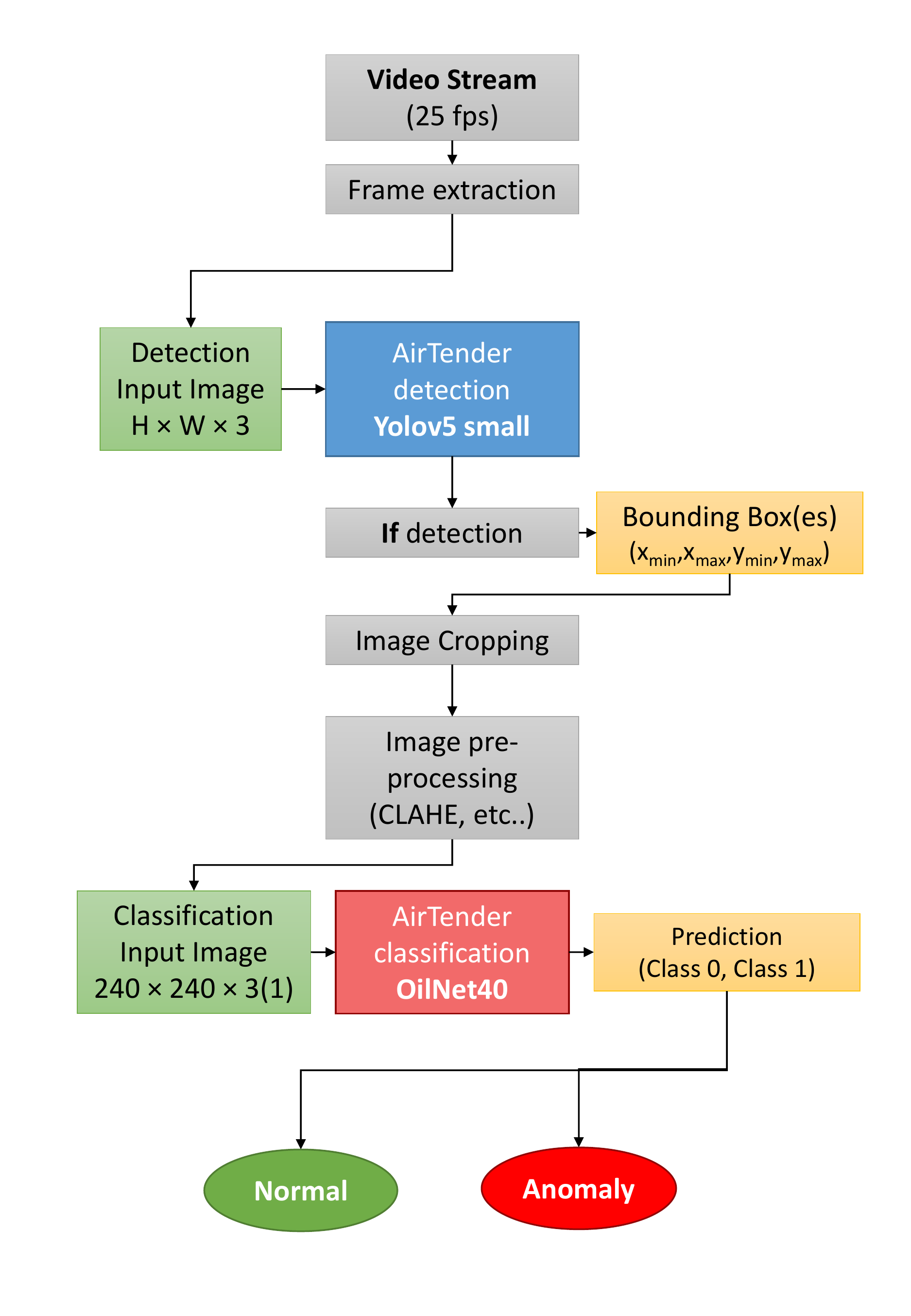}
	\caption{Summary of the key processes to spot oil leaks.}
	\label{fig:workflow}
\end{figure}

The network should be fed the resulting image of the object detection. At a practical level, this is done by cutting out what is inside the bounding box of the object detection. The classifier, in turn, will then assign a class to the AirTender: Normal or Anomaly, according to the status of Airtender. See Figure~\ref{fig:workflow}.

\begin{figure*}[h!]
\centering
\captionsetup{width=.42\linewidth}
\subfloat[Training and validation accuracy for OilNet40 trained on four different versions of the dataset.  The model trained on the dataset preprocessed with CLAHE shows more stability and reaches the highest accuracy. ]
{\includegraphics[width=6.9cm]{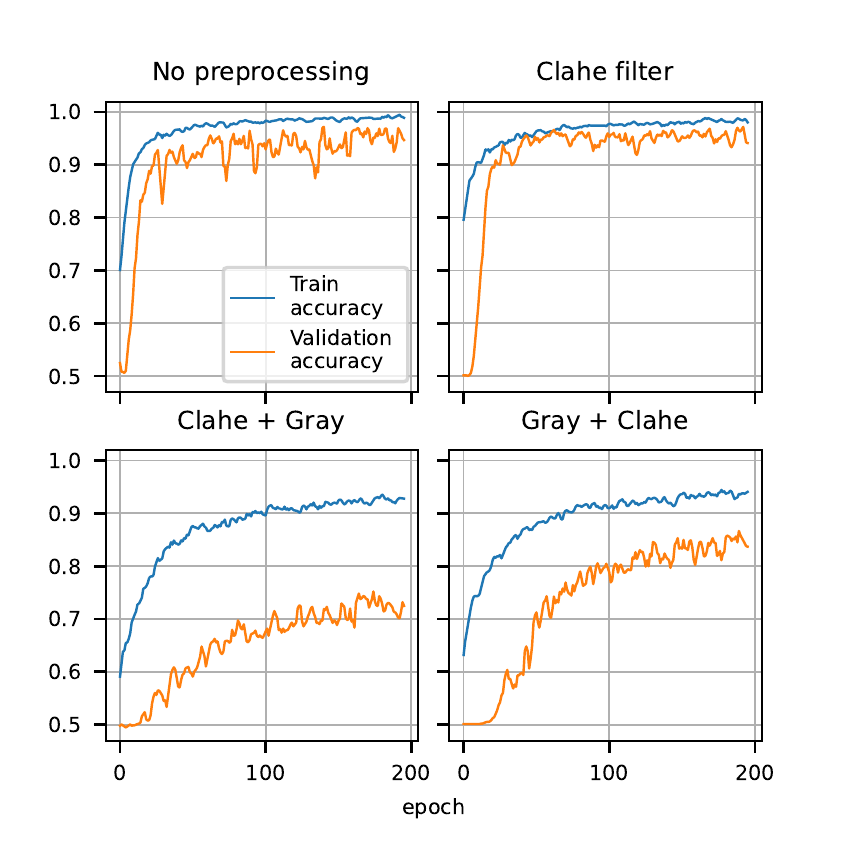}
\label{acc}}
\hfil
\subfloat[Precision and Recall for training and validation of OilNet40. The model performs best on color images.  The model trained on the dataset filtered with CLAHE has the smallest number of false positive and negatives.]
{\includegraphics[width=6.9cm]{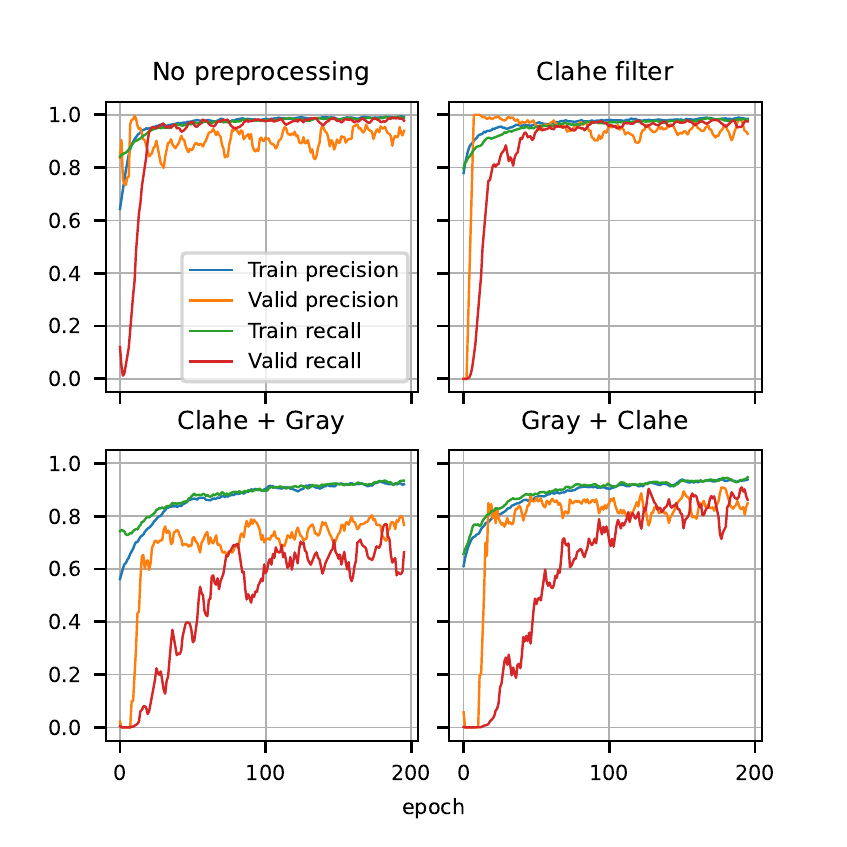}
\label{prec_rec}}
\caption{}
\label{fig:stat}
\end{figure*}

As mentioned above, we trained OilNet40 on four different versions of our original dataset. The results of the training are shown in Figure~\ref{fig:stat}a,b.  In each case, we plot the accuracy of prediction as well as Precision and Recall versus the  number of epochs, for both training and validation.  All metrics (Accuracy on the left plot and Precision and Recall on the right plot) are plotted on the $y$ axis. Notice that we have always split the dataset in 50$\%$ leaks and $50 \%$ 
non-leaks, thus a randomly guessing algorithm is expected to give the correct answer $50 \%$ of the time, while a well-trained Neural Network is expected to give results with confidence well above that threshold. 

From the plots in Figure~\ref{fig:stat}a,b, we see that all metrics improve with the number of epochs, and the better performing algorithm is that trained on images where the CLAHE filter was applied (see Section \ref{image preprocessing}).  The Network trained on grayscale images does not seem to perform as well, reaching a level of accuracy of about $80\%$ in both cases (CLAHE + Gray and Gray + CLAHE).  The grayscale datasets were introduced with the goal of having a more universal algorithm, working independently of the color of AirTender and emitted fluorescence.  The main reason for which we are not able to train an algorithm that generalizes well is presumably due to a small dataset. A larger and more representative dataset is expected to lead to a generalizing algorithm also for the grayscale images of~AirTender. 

\begin{figure}
	\centering
	\hspace{-5pt}\includegraphics[width=7.2cm]{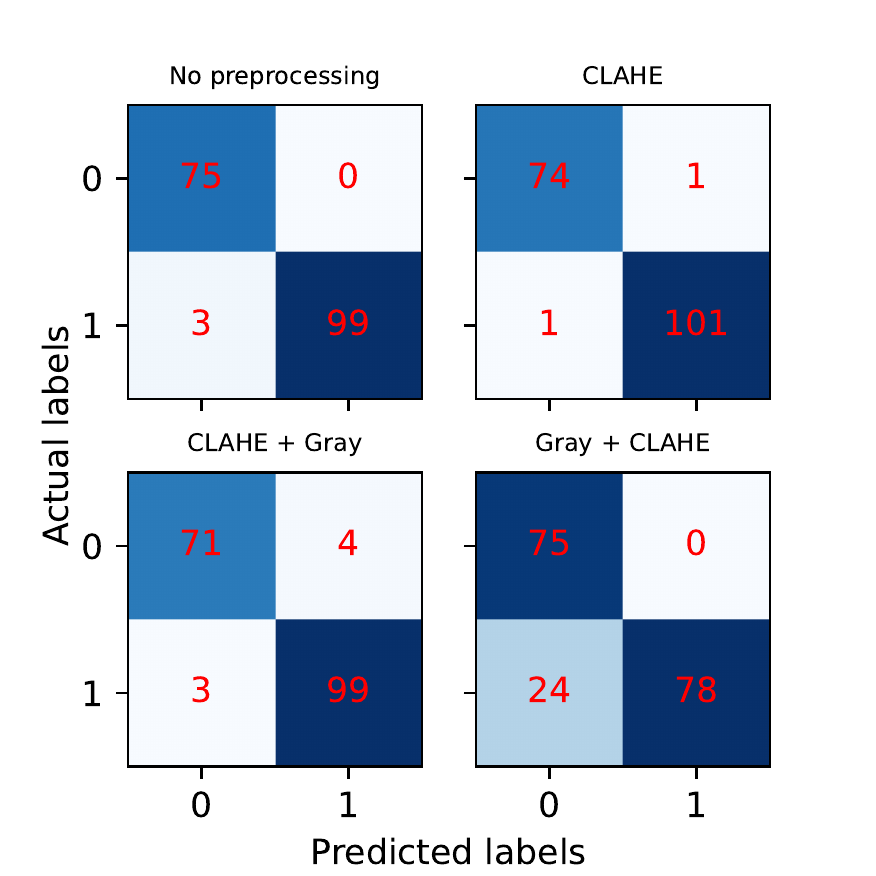}
	\caption{Confusion Matrices for OilNet40 tested on four different versions of the test dataset. The outcomes confirm the results of the training for the color images. In the case of grayscale images, OilNet40 trained on CLAHE + gray dataset seems to perform better than gray + CLAHE even though the training suggested the opposite.}
	\label{fig:confusion_matrix}
\end{figure}

Besides training and validation, we tested our network on a separate test set, for which we plotted confusion matrices. See Figure~\ref{fig:confusion_matrix}.  As mentioned above, the test data are never introduced when training the OilNet40 and correspond to images taken independently of the training and validation sets and at a different time of the day. On the vertical axis,
 we have the actual labels, normal or anomaly (i.e., non-leak or leak image). We can see that the models performing best are those trained on images with no preprocessing or on images where CLAHE was applied, as expected. In particular, in the first case (no preprocessing),
  we have zero false positives but 3 false negatives, while in the second case (CLAHE),
   we have one false positive and one false negative.  In the case of grayscale images, OilNet40 trained on CLAHE + gray dataset seems to perform better than the same model trained on the gray + CLAHE dataset, even though the training suggested the opposite. In the former case, OilNet40 achieves Precision 0.96 and Recall 0.97, while in the latter, it achieves Precision 1 and Recall 0.77, finding more normal configurations than the actual ones.

It is worth noting at this point that, differently from other architectures, CNNs are not quite black boxes as one might think. In fact, in CNNs,
 it is possible to visualize the feature maps, i.e., what a CNN sees after each convolution. This step turns out to be useful to understand whether the filters highlight the regions of interests for a classifier such as
  ours. This gives a view into how an input is decomposed in the classification process.

\begin{figure} 
	\centering
	\includegraphics[width=8.1cm]{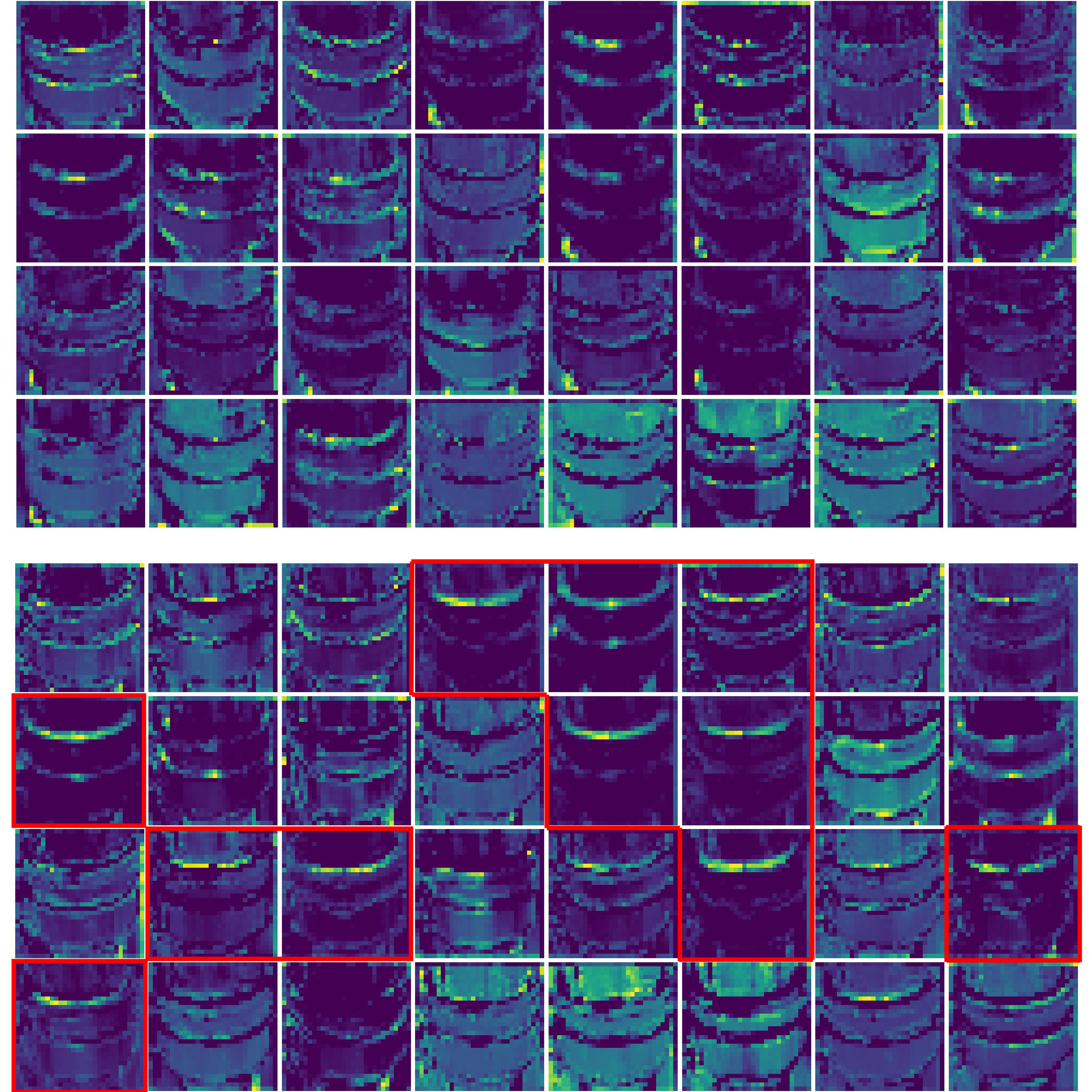}
	\caption{The 32 channels of the third convolutional layer activation on both a normally functioning AirTender (\textbf{above}) and a faulty AirTender (\textbf{below}). In the image below, red contours highlight the feature maps related to the filters which mainly react to the presence of spilled oil on the surface of AirTender core. The same filters are not activated on a similar image of AirTender with no spilled oil, as shown in the upper image.}
	\label{fig:activations}
\end{figure}

In Figure~\ref{fig:activations},
 we show every channel (32 filters in total) of the third layer activation on both a normally functioning AirTender (above) and a faulty AirTender (below).  In Figure~\ref{fig:activations}, image below, red contours highlight the feature maps related to the filters which mainly react to the presence of spilled oil on the surface of AirTender core. The same filters are not activated on a similar image of AirTender with no spilled oil, as shown in the upper image.

At this point, it could be  nice to test the robustness of our results, making the two networks (Yolo and OilNet40) work sequentially, one after the other. Let us see how this is the case.

\subsection{Robustness of the Results}

In order to check the robustness of our results, we tested the complete algorithm, illustrated in Figure~\ref{fig:workflow}, on a live video stream. We installed the TF Lite version of the trained networks on the OrangePi board, to check its actual performance on such a pipeline. We set up a python script following the flux diagram in Figure~\ref{fig:workflow}, to have a working algorithm which first detects AirTender from a live camera and then classifies the image as leak/non-leak. 

To record the video, the AirTender was mounted on a pedestal together with the whole suspension system (spring, damper, etc.) and placed indoor in varying light conditions. To test the classification of anomalous functioning, the AirTender was stained with oil+dye and enlightened with UV light. A one-minute video stream was fed to the OrangePi. 

We found that it can process video streams with a frequency of 0.25 fps. To check the anomaly-detection performances, we recorded a video of both leak/non-leak events and manually labeled each frame, then fed the video-dataset to the OrangePi. The video-dataset consists
 of 1517 frames, 651 normal and 866 anomalous. It turned out that the embedded system can classify images correctly (accuracy) 94$\%$ of the times. We can consider this a good result for a preliminary study. The same test should be run, in the future, on a real~motorcycle.

\section{Concluding Remarks}\label{conclusions}

Our work illustrates a new and efficient way of using on-board devices to detect fault events of a modern suspension system. In particular, we relied on deep learning techniques to detect and classify the status of a special type of motorbike suspension system. The procedure is fast, flexible and easy to improve. 

The algorithm presented in this paper is shown to be efficient on images and video-frames of AirTender that make up our dataset, and consists of two steps. We first focus on the detection of AirTender in the motorbike suspension system. This step turns out to be necessary because a camera placed next to the chassis system of a generic motorbike is not expected to frame AirTender only. It will, in general, frame AirTender and its surrounding environment which is different from motorbike to motorbike.  The second step of our algorithm consists in classifying AirTender as normally functioning or not: a binary classifier is fed with the results of the detection, i.e., the core of AirTender. 

As explained in the text, a broken AirTender is very likely to spill oil. In order to make oil leaks more evident, we set up a small circuit where UV LEDs are pointed at the core of AirTender and are supposed to make the oil fluoresce. All this is amplified by diluting the mechanical oil with fluorescent dye. 

Our work can be easily extended by further enlarging the dataset in order to represent the diversity of leaks that can occur when AirTender breaks. This would allow to train an algorithm that better generalizes to real cases. 

The accuracy of our algorithm depends on a number of factors. First, we considered images of AirTender with a decent resolution. An algorithm operating on very low quality images or videos of Airtender can have poor performance.  
In addition,
 if the oil leak is not visible enough, or if the emitted fluorescence is not of sufficient intensity, the algorithm can misclassify the status of AirTender. To this end, it is necessary to have the UV LEDs directed towards the leaking area.  We hope to improve our algorithm so to include more general setups in future studies.

Finally, we would like to make a few comments on other deep learning techniques that could be used for anomaly detection. To spot oil leaks, we relied on binary classification using a home-made Convolutional Neural Network. It would be interesting to explore other architectures as well. In particular, Convolutional Autoencoders could be employed to spot oil leaks 
for conditions ``far from the normality''. 
In addition,
 oil leaks could be detected by means of semantic segmentation.  For example,  a network such as the U-Net~\cite{ronneberger2015u}, introduced for biomedical image segmentation,
  is sometimes also used  to spot oil leaks on metallic surfaces~\cite{lu2019oil}. 
  We will attempt to explore all these different architectures in future~research. 

\section*{Declaration of competing interests}

No competing interests.

\section*{Acknowledgments}

The study presented in this paper is part of the INNOSUSP project financed to Umbria Kinetics S.r.l. by INVITALIA (IT) Brevetti+.

\IEEEtriggeratref{60}
\bibliography{Bibliography}

\end{document}